\documentclass[sigconf, screen]{acmart}
\usepackage{multirow}
\AtBeginDocument{%
  }

\setcopyright{acmlicensed}
\copyrightyear{2018}
\acmYear{2018}
\acmDOI{XXXXXXX.XXXXXXX}
\acmConference[Conference acronym 'XX]{Make sure to enter the correct
  conference title from your rights confirmation email}{June 03--05,
  2018}{Woodstock, NY}
\acmISBN{978-1-4503-XXXX-X/2018/06}

\begin{document}

\title[StgcDiff]{StgcDiff: Spatial-Temporal Graph Condition Diffusion for \\ Sign Language Transition Generation}

\author{
    Jiashu He$^{1}$, Jiayi He$^{1}$, Shengeng Tang$^{1}$*, Huixia Ben$^{2}$, Lechao Cheng$^{1}$, Richang Hong$^{1}$
}

\affiliation{%
  \institution{$^{1}$Hefei University of Technology, $^{2}$Anhui University of Science and Technology}
   \city{Hefei}
  \country{China}
}


\begin{abstract}
Sign language transition generation seeks to convert discrete sign language segments into continuous sign videos by synthesizing smooth transitions. However, most existing methods merely concatenate isolated signs, resulting in poor visual coherence and semantic accuracy in the generated videos. Unlike textual languages, sign language is inherently rich in spatial-temporal cues, making it more complex to model. To address this, we propose StgcDiff, a graph-based conditional diffusion framework that generates smooth transitions between discrete signs by capturing the unique spatial-temporal dependencies of sign language. Specifically, we first train an encoder-decoder architecture to learn a structure-aware representation of spatial-temporal skeleton sequences. Next, we optimize a diffusion denoiser conditioned on the representations learned by the pre-trained encoder, which is tasked with predicting transition frames from noise. Additionally, we design the Sign-GCN module as the key component in our framework, which effectively models the spatial-temporal features. Extensive experiments conducted on the PHOENIX14T, USTC-CSL100, and USTC-SLR500 datasets demonstrate the superior performance of our method.
\end{abstract}

\keywords{Do, Not, Us, This, Code, Put, the, Correct, Terms, for, Your, Paper}


\maketitle

\begin{figure}
    \centering
    \includegraphics[width=1\linewidth]{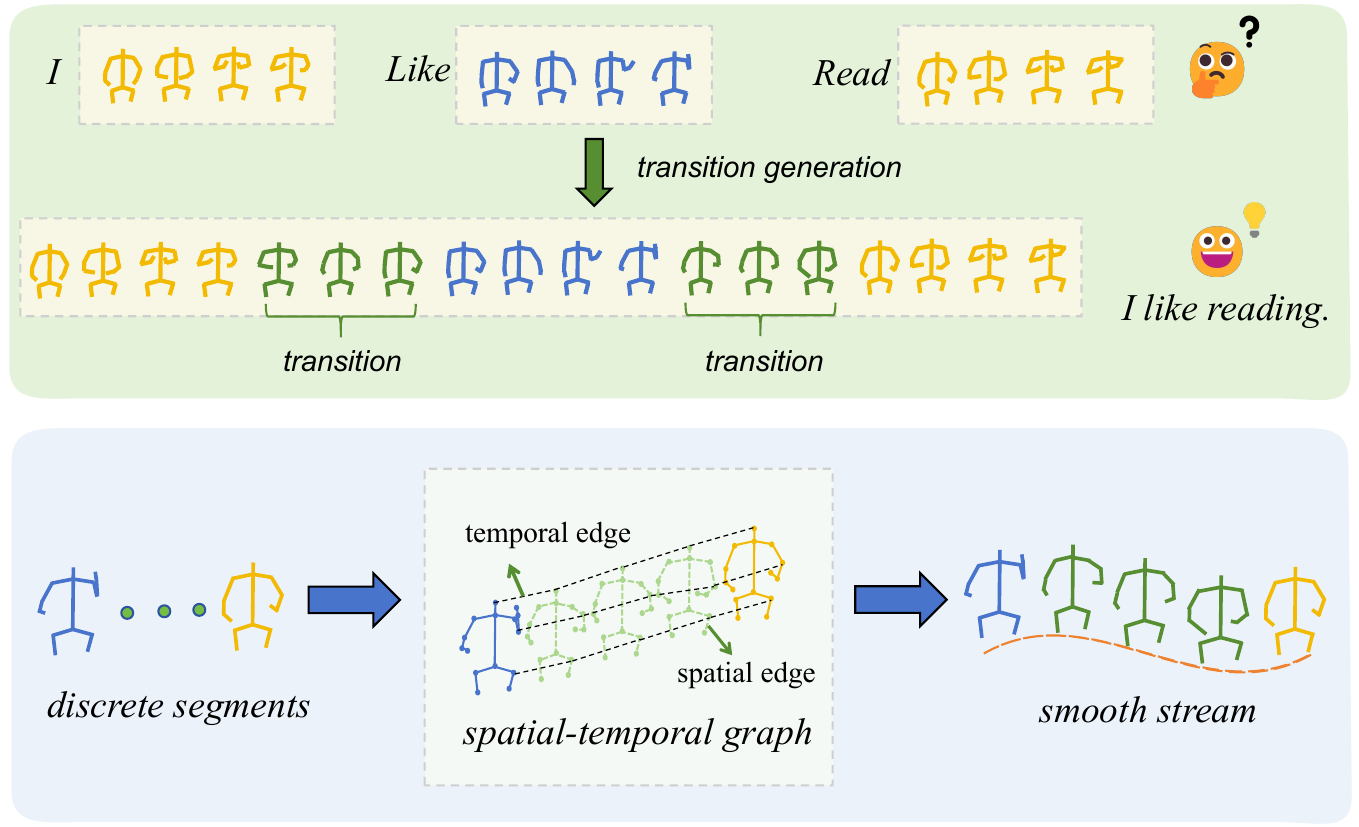}
    \vspace{-3mm}
    \caption{
\textbf{Top}: An illustration of the sign language transition task, where transitions (green) are generated between three discrete sign segments (blue and yellow) to produce a continuous sign language video. \textbf{Bottom}: The proposed spatial-temporal graph modeling framework, where temporal and spatial dependencies are jointly leveraged to synthesize smooth transitions between discrete segments.
}
    \label{fig:task}
    \vspace{-3mm}
\end{figure}

\section{Introduction}
Sign language is the primary mode of communication for the deaf and hard-of-hearing communities, encompassing rich spatial and temporal information. Although significant advancements have been made in sign language research~\cite{tang2024sign, jiang2021skeleton, tang2024discrete}, the field still faces challenges, particularly due to the limited availability of large-scale sentence-level datasets. As illustrated at the top of Figure~\ref{fig:task}, a potential solution is to synthesize continuous sign language from isolated sign datasets~\cite{huang2018attention, chai2014devisign, dreuw2008benchmark, li2020word, zhang2016chinese}. However, most previous approaches~\cite{wang2002method, sagawa2002teaching} simply splice isolated signs together to form sentences. This method results in unnatural transitions and abrupt motion shifts. Unlike textual language, which primarily relies on the linear arrangement of words, sign language requires smooth transitions between signs to maintain fluency and naturalness while conveying the intended meaning. Without effective transition modeling, directly concatenating individual signs often leads to awkward interruptions or abrupt shifts, making the synthesized sentence difficult to understand. As such, the challenge of seamlessly connecting individual signs to form continuous, meaningful sign language sentences has become a focal point of recent efforts.

Existing works on sign language transition generation~\cite{cruz2024generative, tang2024discrete} primarily focus on achieving smooth transitions, often overlooking the spatial-temporal cues inherent in signs. Specifically, these approaches tend to treat transitions as purely temporal interpolation problems, without explicitly capturing the spatial structure of human joints. As a result, this can lead to overly smooth motion or physically implausible joint movements, especially when there is a significant motion gap between consecutive segments or when complex spatial articulation is involved. Natural transitions between discrete sign language segments rely not only on smooth temporal connections but also on maintaining the continuity of the spatial structure. To ensure the naturalness and physical plausibility of transition frames, spatial-temporal modeling becomes crucial. By capturing both the spatial and temporal dependencies in sign language movements, we can better account for the interactions between joints and the continuous flow of motion. This calls for a spatial-temporal representation that accurately conveys both the spatial positions of the joints and their dynamic changes over time, thereby generating more natural and coherent transition frames.

To address this issue, StgcDiff is proposed, a diffusion model designed to generate smooth transitions between discrete sign language segments by effectively capturing the inherent spatial-temporal cues. Specifically, Sign-GCN is introduced, combining a spatial structure extraction module with a temporal feature perception module. A GCN-based encoder-decoder framework is pre-trained, enabling more accurate and detailed reconstructions compared to previous methods. By extracting structure-aware features from the observed frames through the encoder, conditional guidance is provided to the diffusion model, allowing it to generate transitions that preserve the spatial integrity of sign poses while ensuring smooth and coherent temporal motion. The main contributions of this work are summarized as follows:
\begin{itemize}
    \item A diffusion framework based on spatial-temporal graphs is proposed for sign language transition generation. Additionally, a pre-trained encoder-decoder architecture is employed to learn structure-aware representations of observed sign segments, which serve as conditional information to guide the generation of transition frames. 
    \item We design a Sign-GCN module to effectively model the spatial-temporal dependencies in sign pose sequences. In the spatial domain, a GCN is leveraged to capture the relationships between different joints in the sign language skeleton. In the temporal domain, a multi-scale TCN is employed to capture the long-range temporal dependencies, allowing the module to effectively perceive the evolution of the sign language sequence over time. 
    \item Extensive experiments are conducted on three publicly available sign benchmarks, PHOENIX14T\cite{Camgoz_Hadfield_Koller_Ney_Bowden_2018}, USTC-CSL100\cite{huang2018video}, and USTC-SLR500~\cite{huang2018attention}, demonstrating the superior performance of the proposed method in generating high-quality transitions, surpassing existing approaches. 
\end{itemize}

\section{Related Work}
\subsection{Sign Language Production \& Transition}
Sign language is the main form of communication for deaf and hard-of-hearing people. Current works focus on building a two-way conversion system from sign language to text. Despite significant progress in Sign Language Translation (SLT)~\cite{gong2024llms, zhao2024conditional, chen2022simple, tang2021graph} and Recognition (SLR)~\cite{zhao2024masa, Koller_2020, zhao2023best, zuo2023natural}, substantial challenges remain in Sign Language Production (SLP). 

Sign language production aims to convert text into real sign language videos or pose sequences. Existing SLP works can be divided into rule-based and data-driven methods. Rule-based methods~\cite{joy2020developing, de2023querying, aliwy2021development} rely on predefined sign language dictionaries to map text to corresponding signs. However, the construction of sign language dictionaries entails not only substantial costs but also specialized expertise in sign linguistics. With the rapid development of deep learning, data-driven methods have become the mainstream research paradigm in the field of SLP. Saunders \emph{et al.}~\cite{saunders2020progressive} introduces Progressive Transformers for SLP, which utilize a Transformer-based architecture to enhance the fluency and accuracy of generated sign language. Recently, diffusion-based methods ~\cite{xie2024g2p, tang2024GCDM, tang2024sign} have been applied to SLP, leveraging the unique properties of diffusion models to generate sign pose sequences under semantic guidance. 

Despite significant progress, limited long-sequence sign language data remains the key bottleneck for sign language. To address this issue, Zeng \emph{et al.} ~\cite{zeng2020highly} develops an adaptive interpolation algorithm based on kinematic features that enables natural transition generation. Meanwhile, Cruz and Bejarano ~\cite{cruz2024generative} propose a Residual Vector Quantized Variational Autoencoder (RVQ-VAE) that generates intermediate sign pose frames and mitigates unnatural transitions from missing frames in sign language production. Recently, Tang \textit{et al.} \cite{tang2024discrete} proposes a method converting the unsupervised transition modeling into a supervised learning task via a random masking strategy, which provides a new direction for transitions. However, these methods neglect the inner-frame kinematic dependencies between body joints, causing a lack of coherence and accuracy. To this end, we propose Sign-GCN to generate smooth transitions between discrete symbols by capturing inherent spatial-temporal information.

\begin{figure*}[tbh]
  \centering
  \includegraphics[width=1.0\textwidth]{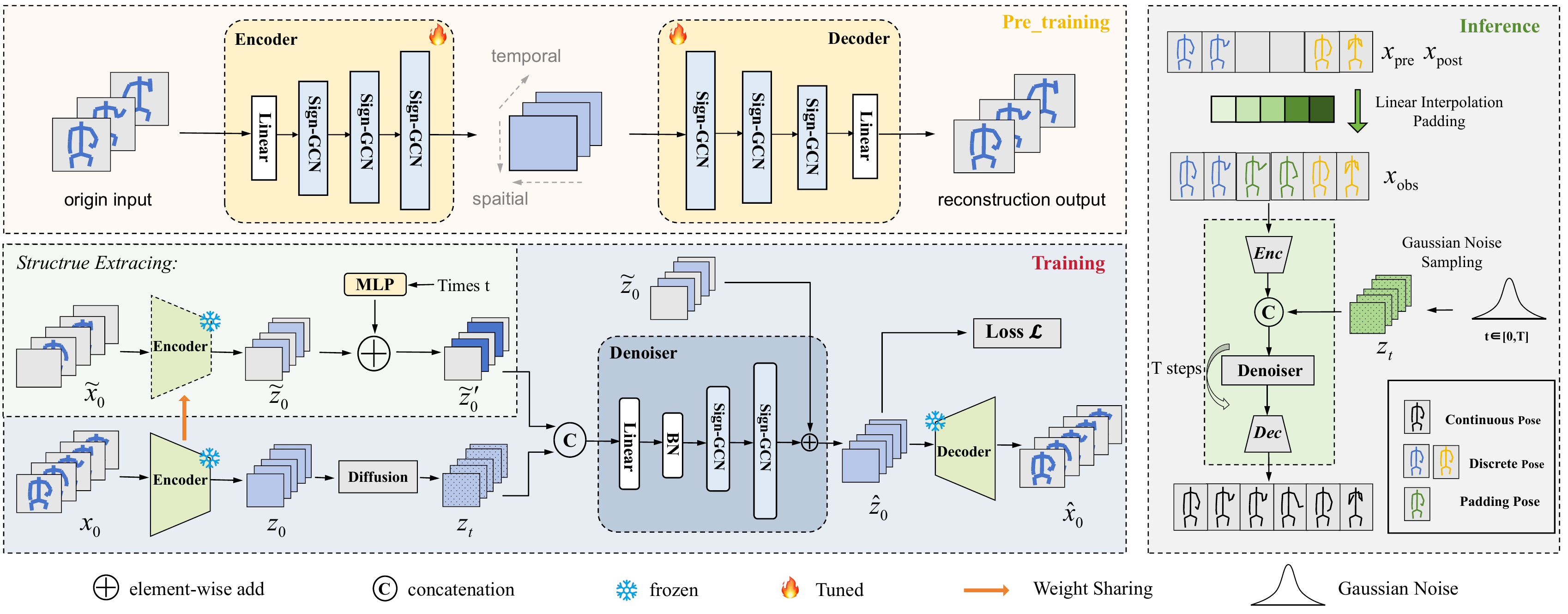}
  \caption{The overview of the proposed method. In the pre-training phase (left top), a long-duration sign video is processed by our gragh-based encoder-decoder architecture to reconstruct the input, which aims to learn the spatial-temporal representation. In the training phase (left bottom), we train a conditional diffusion model to generate smooth transitions between discrete segments. We sample the noisy sign skeleton features $z_t$ following the diffusion process. The diffusion denoiser predicts the feature of transitions from $z_t$ guided by the learned representation extracted from the randomly masked input video $\tilde{x}_0$. For optimization, MAE loss $\mathcal{L}_{MAE}$  is utilized in the training stage. In the inference phase (right), we apply the linear interpolation padding to initialize transition frames aiming to connect discrete segments. Then, our diffusion model (green part)  refines these transitions by iteratively denoising $T$ steps from gaussian noise, generating a continuous sign video coherent with both context segments. 
  }
  \label{fig:framework}
\end{figure*}

\subsection{Diffusion-based Generative Models}
Diffusion models~\cite{Sohl-Dickstein_Weiss_Maheswaranathan_Ganguli_2015, Ho_Jain_Abbeel_Berkeley} have become a popular generative paradigm due to their ability to produce high-quality and diverse samples by modeling the generation process as the reversal of a gradual noise-adding process. Initially applied in image generation, diffusion models have shown remarkable performance in generating complex structures across multiple domains, including image generation~\cite{rombach2022high}, video synthesis~\cite{yang2024fresco, lee2024grid}, 3D shape modeling~\cite{vahdat2022lion}, and human motion generation~\cite{zhang2024motiondiffuse,zhou2024emdm}.

Despite their success in vision and motion tasks, the application of diffusion models in Sign Language Production (SLP) remains relatively underexplored. Recent efforts such as G2P-DDM~\cite{xie2024g2p} introduce discrete diffusion techniques to map gloss sequences to pose sequences, while VQ-GCDM~\cite{tang2024GCDM} employs a gloss-conditioned multi-hypothesis diffusion framework for diverse sign generation. However, these approaches primarily focus on the semantic correctness of generated poses from linguistic inputs and overlook the challenge of maintaining temporal continuity, especially across discrete sign segments. To this end,  Tang \textit{et al.} \cite{tang2024discrete} proposes Sign-D2C, a conditional diffusion model to synthesize contextually smooth transition frames, enabling the seamless construction of continuous sign language sequences.

Our work complements these studies by leveraging a spatial-temporal graph conditional diffusion framework for generating smooth transition frames. We focus on modeling the spatial and temporal relationships of sign poses, a crucial aspect for constructing coherent and continuous sign language videos.

\subsection{Graph Convolutional Networks}
Graph Convolutional Networks (GCNs)~\cite{kipf2016semi} have been widely applied in many applications like visual recognition~\cite{hua2013collaborative, hong2020graph, hu2021novel, hu2024crd, long2015multi}, object detection~\cite{wang2021joint, islam2020doa}, action localization~\cite{zeng2019graph, islam2021hybrid}, trajectory prediction~\cite{shi2021sgcn}, and image captioning \cite{dong2021dual}. Many advances~\cite{liu2020disentangling, chen2021channel, shi2019two, tang2018deep, ye2020dynamic} have demonstrated the potential of GCNs in modeling the complex spatial-temporal patterns inherent in skeletons. Among these methods, Yan \emph{et al.}~\cite{yan2018spatial} proposes Spatial-Temporal Graph Convolutional Networks (ST-GCNs), which first demonstrate the advantages of GCNs in spatial-temporal modeling for human action recognition.

Motivated by these advances in action recognition, GCNs have also been widely adopted in sign language, such as Sign Language Recognition (SLR)~\cite{tunga2021pose, jiang2021skeleton, vazquez2021isolated, guo2023interactive, kan2022sign, arib2025signformer}. For instance, Jiang \emph{et al.}~\cite{jiang2021skeleton} introduces SL-GCN and SSTCN within a multi-modal framework to enhance isolated SLR accuracy. Tunga \textit{et al.} \cite{tunga2021pose} proposes a model that combines GCN and BERT to effectively capture spatial and temporal features from skeleton sequences for pose-based sign language recognition. Recent studies such as Parelli \textit{et al.} \cite{parelli2022spatio} and Arib \textit{et al.}~\cite{arib2025signformer} propose incorporating GCNs into continuous recognition and translation pipelines. Compared to traditional methods, GCNs offer superior representational power for feature extraction and modeling of sign language skeleton sequences. While previous works have leveraged GCNs primarily for recognition and translation tasks, our method focuses on sign language generation, specifically addressing smooth spatial-temporal transitions between discrete sign segments, a relatively unexplored aspect in GCN-based Sign Language Production (SLP) \cite{yu2022graph}.  

Inspired by Spatial-Temporal Graph Convolutional networks~\cite{yan2018spatial}, we propose a graph-based sign language transition. Our method eliminates the previous neglect of local skeletal structural consistency and models sign poses in a spatial-temporal skeleton graph with high granularity, thereby improving the plausibility and coherence of the generated transition frames. 


\section{Preliminaries: Unsupervised to Supervised}
Generating smooth transitions between discrete sign segments is challenging due to the lack of ground-truth transition data, as real-world sign datasets often lack explicitly annotated transitions.

To address this, we follow the core idea proposed in Sign-D2C~\cite{tang2024discrete} and transform the problem into a supervised learning task through a random masking strategy. Specifically, during training, long continuous sign language sequences are treated as complete ground-truth trajectories. We randomly mask out short temporal segments within these sequences, effectively simulating missing transitions between adjacent observed segments. This masked data setup creates pseudo-gaps similar to those found when stitching together isolated signs in practice.
 
Given a complete sign language sequence $X = \{x_1, x_2, \dots, x_T\}$ with T frames, and each $x_t$ represents the 3D coordinates of $N$ joints at time $t$. We define a masking set $\mathcal{M} \subset \{1, 2, \dots, T\}$ containing randomly selected time indices. The observed frames are:
\begin{equation}
    X_{obs} = \{x_t \mid t \notin \mathcal{M}\}, \quad X_{mask} = \{x_t \mid t \in \mathcal{M}\}
\end{equation}

The model is then trained to predict the masked frames $X_{mask}$ based on the surrounding observable context $X_{obs}$, thereby learning to reconstruct smooth, temporally consistent transitions in a supervised manner.

By reformulating the original unsupervised task as a supervised reconstruction problem, the model benefits from standard training objectives and gains the ability to generate naturalistic transitions conditioned on surrounding sign poses.

\section{Methodology}
In this work, we propose a graph-based framework for generating smooth transitions between discrete sign segments to create continuous sign language videos. This section will outline the sign pose modeling (\hyperref[sec:sign pose]{Sec. 4.1}), followed by  the perceptual structure representation (\hyperref[sec:pretraining]{Sec. 4.2}), the overall framework (\hyperref[sec:framework]{Sec. 4.3}) and the Sign-GCN module (\hyperref[sec:Sign-GCN]{Sec. 4.4}).

\subsection{Spatial-Temporal Modeling of Sign Pose}
\label{sec:sign pose}
Let $X \in \mathbb{R}^{C \times T \times V}$ denote the input sign pose sequence, where $C$ is the input channel ($C=3$ for 3D coordinates), $T$ is the number of frames, and $V$ is the number of joints. To model both spatial and temporal dependencies for the sign language transition task, we construct a spatial-temporal graph $\mathcal{G} = (\mathcal{V}, \mathcal{E})$. Each node $v_i^t \in \mathcal{V}$ corresponds to joint $i$ at time $t$, and the edge set $\mathcal{E}$ is defined as:

\begin{equation}
\mathcal{E} = \underbrace{\{(v_i^t, v_j^t)\}}_{\text{Spatial edges}} \cup \underbrace{\{(v_i^t, v_i^{t+1})\}}_{\text{Temporal edges}},
\end{equation}
where the spatial edge $(v_i^t, v_j^t)$ denotes joints $i$ and $j$ are physically connected at time $t$, and the temporal edge $(v_i^t, v_i^{t+1})$ denotes the same joint $i$ across consecutive time steps $t$ and $t+1$. 

The spatial edges model the natural anatomical structure of the sign pose, while temporal edges connect the same joint across consecutive frames to capture motion dynamics. By explicitly modeling both spatial and temporal dependencies, the spatial-temporal graph provides a strong structural prior for the diffusion model.

\subsection{Structure-aware Representation Learning}
\label{sec:pretraining}
In this section, we introduce a GCN-based encoder-decoder pre-training architecture and pre-train it prior to training the full diffusion framework, which aims to comprehensively capture sign poses and learn a structure-aware representation of spatial-temporal skeleton sequences. 

\noindent\textbf{Encoder-decoder Architecture}
As shown in Figure~\ref{fig:framework}, our encoder-decoder architecture contains an encoder and decoder, and we adopt Sign-GCN 
as the backbone of the encoder-decoder architecture. The encoder $\mathcal{E}$ consists of three stacked Sign-GCN layers that map sign poses to the latent space with rich spatial-temporal information. The decoder $\mathcal{D}$ mirrors the structure of the encoder with three Sign-GCN layers, which reconstruct the hidden embedding under the supervision of the reconstruction criterion. 

Given a sign pose sequence $\mathcal{X} \in \mathbb{R}^{3 \times T \times V}$  as the input of our encoder-decoder architecture, we firstly embed all skeleton joints and their topology into the spatial-temporal graph $\mathcal{G}$. Then, $\mathcal{X} \in R^{3 \times T \times V}$ is linearly transformed to $\mathcal{X}^{\prime} \in R^{C \times T \times V}$ with learnable parameters.  In practice, we set C to 8. With the Encoder $\mathcal{E}$  based on Sign-GCN, we gradually project the input into higher-dimensional spaces with dimensions 16, 64, and 128, resulting in a latent representation $Z \in \mathbb{R}^{128 \times T \times V}$. The decoder progressively reduces the latent feature dimension from 128 to 64, then to 16, and finally reconstructs the input sequence back into the original pose domain as $\hat{\mathcal{X}} \in \mathbb{R}^{3 \times T \times V}$ with a trainable linear layer. Moreover, we employ batch normalization for $\mathcal{X}^{\prime}$ in the encoder. The process of our reconstruction is formalized as:
\begin{equation}
\left\{
\begin{aligned}
Z &= \mathcal{E}(\mathcal{X}), & Z  &\in \mathbb {R}^{128 \times T \times V}, \\
\hat{\mathcal{X}} &= \mathcal{D}(Z), & \hat{\mathcal{X}} &\in \mathbb{R}^{3 \times T \times V}
\end{aligned}
\right.
\end{equation}
The objective of pre-training can be formalized as minimizing the divergence between $\mathcal{X}$ and $\hat{\mathcal{X}}$.

\noindent\textbf{Reconstruction Criterion}
In this paper, we optimize the encoder-decoder architecture using the Mean Absolute Error (MAE) loss to constrain the consistency of the reconstructed output \( \mathcal{\hat X} = \{ \mathbf{\hat x}_{t,v} \}_{t=1, v=1}^{T,  V} \) and the input pose sequence \( \mathcal{X} = \{ \mathbf{x}_{t,v} \}_{t=1, v=1}^{T,V} \):
\begin{equation}
\mathcal{L}_{\text{rst}} = \frac{1}{TV} \sum_{t=1}^{T} \sum_{v=1}^{V}  \left| x_{t,v} - \hat{x}_{t,v} \right|
\end{equation}

By training the GCN-based encoder-decoder architecture to reconstruct the sign pose sequence,  the pre-trained network can comprehensively perceive the skeleton structure and obtain spatial-temporal representation.

\subsection{StgcDiff Framework}
\label{sec:framework}
In this section, we describe the proposed conditional diffusion framework StgcDiff, as a unified framework for sign language transition. Figure~\ref{fig:framework} illustrates the overall pipeline of our method. 

Concretely, we optimize a Sign-GCN denoiser to generate smooth transition frames by training a conditional diffusion model. With the aforementioned pre-training, the encoder learns to extract structure-aware representation applicable for sign language transition tasks. Meanwhile, this structure consistency of the encoder and the denoiser also helps maintain the consistency of the spatial-temporal features of graph data.

\noindent\textbf{Conditioning.} 
Given the sign video dataset $p_{video}$, we take a $T$-frame original video denoted as $x_{0} \in \mathbb{R}^{3 \times T \times V}$, where $x_{0}$ follows the distribution  $p_{video}$. Following ~\cite{tang2024discrete},  we randomly mask out the original video $x_{0}$ to simulate the absence of transition frames, resulting in a masked sign video $\tilde{x}_0 \in \mathbb{R}^{ 3 \times T \times V}$ . Then,  $\tilde{x}_0$ is encoded by the pretrained GCN-encoder $\mathcal{E}$ as $\tilde{z}_0 \in \mathbb{R}^{C \times T \times V}$, where $C$, $T$ and $V$ indicate the channels, the frame number and the joint number. 
 Meanwhile, we employ a MLP for the times of adding noise $t$  and  the time embedding is added to $\tilde{z}_0$ to incorporate temporal information:
\begin{equation}
\tilde{z}^{\prime}_0 = \tilde{z}_0 + \text{MLP}(t),
\end{equation}
where $\tilde{z}^{\prime}_0$ denotes the conditional guidance for the generation of sign language transition frames. 

\noindent\textbf{Denoiser Architecture.}  
Our denoising module takes as input the noised transition frames along with context information and time embedding, which reconstructs the original representation by iteratively removing the noise $\epsilon$. 
First, we apply a linear transformation to project the input into a new feature space of  $\mathbb{R}^{F' \times T \times V }$.
Then,  we put the feature map to two Sign-GCNs, which is responsible for extracting features in dimensions 64 and 128 respectively. To do this, we can obtain a new feature. In additional to further ensure the generated transitions are guided by context information, we also inject the masked encoder output $\tilde{z}_0$ into the final residual connection, which helps preserve semantic relevance and stabilizes the denoising process. And the effectiveness of incorporating the residual connection at the final stage will be validated in the subsequent ablation study.  

\noindent\textbf{Training Phase.}  
In the training stage, the latent code  $z_{0}$ of  sign video is corrupted by Gaussian noise $\epsilon$ as $z_{t}$ whose size is same as $z_{0} \in \mathbb{R}^{C \times T \times V}$. This process is formulated as follows:
\begin{equation}
z_t = \sqrt{\bar{\alpha_t}} z_0 + \sqrt{1 - \bar{\alpha_t}} \epsilon, \quad \epsilon \sim \mathcal{N}(0, I)
\end{equation}

In order to incorporate with sign observations which provide the necessary context for our model to predict the masked segments accurately, our denoiser takes $z_{t}$ concatenated with the condition $\tilde{z}^{\prime}_0$ as the input:
\begin{equation}
\tilde{z_t} = [z_t,\tilde{z}^{\prime}_0], \quad \tilde{z}_t \in \mathbb{R}^{(2 \times C) \times T \times V}
\end{equation}
where $\tilde{z}_t$ represents the final input to the denoiser. Finally, we utilize the denoiser to gain
$\hat{z}_0$, the final prediction of the same shape as $z_0$.

During the diffusion model training phase, we employ the denoising loss $\mathcal{L}_{denoise}$ to ensure that the denoiser’s predictions $\hat{z}_0$ remain consistent with the original latent features $z_0$ :
\begin{equation}
\mathcal{L}_{denoise} = \mathbb{E}_{z_0,t} \left[ \left| z_0 - \hat{z}_0 \right| \right]
\end{equation}
The denoiser is trained by minimizing the loss $\mathcal{L}_{denoise}$ directly in an end-to-end manner.

\noindent\textbf{Inference Phase.}
At inference time, our goal is to generate smooth intermediate frames between discrete sign segments, resulting in seamless synthesis of continuous sign language videos. The transition segment $x_{\text{trans}}\in \mathbb{R}^{3 \times T_{trans} \times V}$ is initialized using a linear interpolation between the last frame of the previous segment  $x_{\text{pre}}\in \mathbb{R}^{3 \times T_{pre} \times V}$ and the first frame of the next segment  $x_{\text{post}}\in \mathbb{R}^{3 \times T_{post} \times V}$, where $T_{trans}$, $T_{pre}$ and $T_{post}$ respectively represent the number of transition frames, previous observation frames and next observation frames. We aggregate the three segments to obtain a entire video sequence as $x_{obs}$. The above description can be expressed as:
\begin{equation}
x_{obs} = Concat(x_{pre}, x_{trans} , x_{post})
\end{equation}
Then, the encoder extracts a latent representation from $x_{obs}$, formulated as $z_{obs} = \mathcal{E}(x_{obs})$, which provides guidance for the process of denoising. Starting from pure Gaussian noise $z_t$, the model progressively denoises the sequence across $t$ steps conditioned on $z_{obs}$:

\begin{equation}
\hat z_{t-1} = \text{Denoiser}(z_t, t , z_{\text{obs}})
\end{equation}
until the final output $\hat{z}_{\text{0}}$ is obtained. Finally, the decoder maps the final latent representation ${\hat{z}_0}$ back to the pose space by pre-trained decoder $\mathcal{D}$, producing smooth and natural transitions that are coherent with both context segments. 

\begin{figure}
    \centering
    \includegraphics[width=1\linewidth]{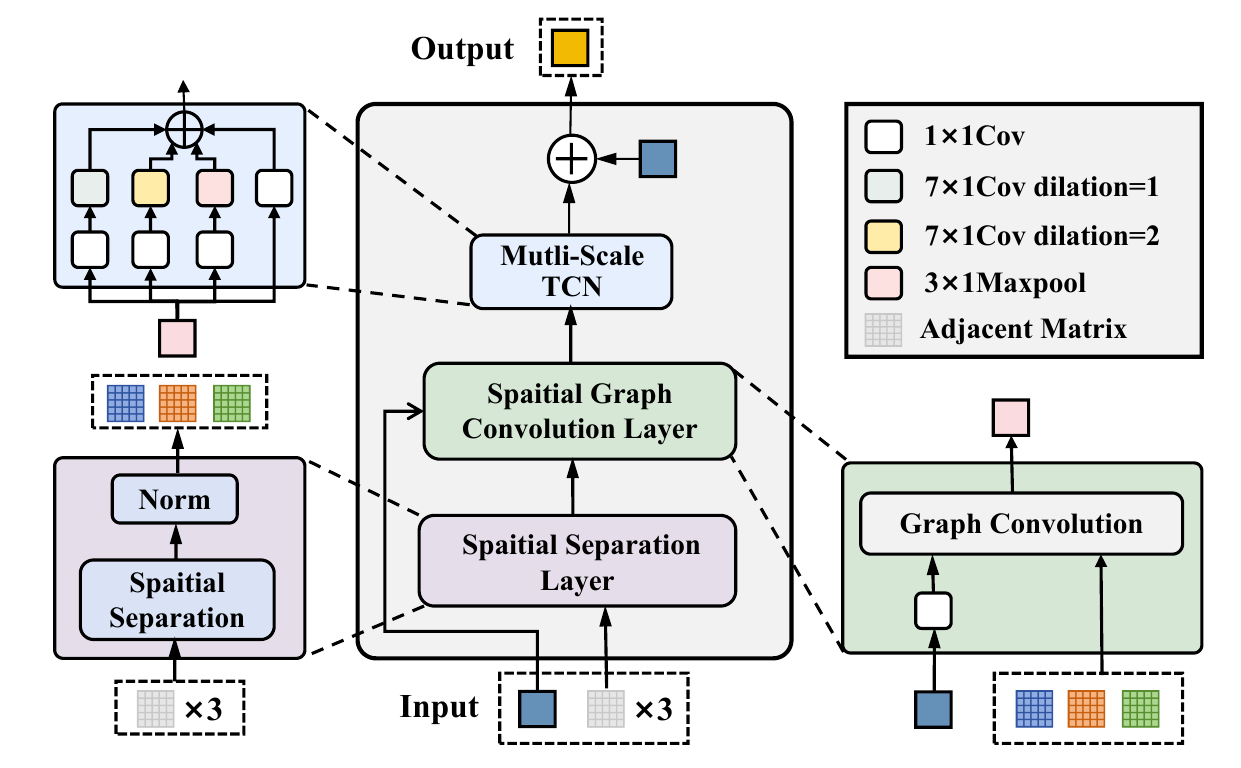}
    \caption{The main components of the Sign-GCN module, include the Spatial Separation Layer, Spatial Graph Convolution Layer, and Multi-Scale Temporal Convolutional Network (TCN), which together model the spatial and temporal dependencies in sign language sequences.}
    \label{fig:Sign-GCN}
\end{figure}

\subsection{Sign-GCN Module}
\label{sec:Sign-GCN}
As shown in Figure~\ref{fig:Sign-GCN}, we design a module named Sign-GCN that integrates spatial and temporal modeling in a unified structure. This module is composed of three main parts: a Spatial Separation Layer for adjacency matrix construction, a spatial graph convolution layer for spatial modeling, and a multi-scale temporal convolutional network for temporal modeling.

Firstly, the adjacency matrices are constructed and normalized based on spatial strategies (\emph{i.e.}, center, centripetal, and centrifugal) introduced in~\cite{yan2018spatial}, resulting in three spatial adjacency matrices ${A_k}{k=1}^{3} \in \mathbb{R}^{3 \times V \times V}$. Next, the feature map $f{\text{in}} \in \mathbb{R}^{C_{\text{in}} \times V \times T}$ is input to the spatial graph convolutional layer, where $C_{\text{in}}$ represents the number of input channels, $V$ is the number of nodes, and $T$ is the time dimension. Following the Graph Convolution formulation proposed by Kipf and Welling~\cite{kipf2016semi}, the output of the graph convolutional layer is computed as:

\begin{equation}
\label{3.1}
f_{\text{out}} = \sum_k^{k_s} ( f_{\text{in}}\cdot A_k ) W_k,
\end{equation}
\begin{equation}
A_k = D_k^{-\frac{1}{2}} (\tilde{A}_k + I) D_k^{-\frac{1}{2}}, \quad D^{ii} = \sum_j (\tilde{A}_k^{ij} + I_{ij}),
\label{3.2}
\end{equation}
where $f_{\text{out}} \in \mathbb{R}^{C_{\text{out}} \times V \times T}$ is the output feature map of the spatial graph convolution layer, $k_s$ denotes the number of spatial kernels ($k_s$=3), $\tilde{A}_k$ is the undirected adjacency matrix representing intra-body connections, $I$ is the identity matrix, and $W_k$ is the trainable weight matrix corresponding to each adjacency matrix.

In the temporal domain, we introduce a Multi-scale TCN to capture temporal features in consecutive frames. As shown in Figure~\ref{fig:Sign-GCN}, the Multi-TCN module contains four parallel branches, each containing a $1 \times 1$ convolution to reduce channel dimension to $C_{out}/4$. Two branches contain two temporal convolutions with different dilation factors $d$ , each of which uses a  $k_t \times 1$  $2D$ convolution operating on (T, V), where $k_t$ is the temporal kernel size. And one branch contains a $3 \times 1$ MaxPool layer following $1 \times 1$ convolution. Finally, the outputs from all four branches are concatenated along the channel dimension and summed with the residual connection to obtain the final output $y$. Specifically, the residual is derived from the input feature map $f_{in}$, either directly or through a $1 \times 1$ convolution when $C_{in}$  is different from $C_{out}$:

\begin{equation}
y = TCN(f_{out}) + W_r \cdot f_{in},
\end{equation}
where $W_r$ denotes the trainable parameters of the residual connection. 

With the  spatial graph convolution layer and the Multi-scale TCN, our Sign-GCN network can extract and accumulate spatial and temporal interactive features simultaneously. 

\section{Experiments}

\subsection{Experimental Settings}
\noindent\textbf{Datasets.}
We evaluate our method on three publicly available sign language datasets: PHOENIX14T~\cite{Camgoz_Hadfield_Koller_Ney_Bowden_2018}, USTC-CSL100~\cite{huang2018video}, and USTC-SLR500~\cite{huang2018attention}. The PHOENIX14T dataset comprises 8,257 videos spanning 2,887 distinct German words and 1,066 glosses, renowned for its inherent complexity. The USTC-CSL100 dataset contains 100 Chinese sign language sentences performed by 50 signers, which we partitioned into 4,000 training instances and 1,000 test instances following specifications in ~\cite{guo2018hierarchical} . The USTC-SLR500 contains 500 isolated Chinese signs performed by 50 signers. Following Huang \emph{et al.}~\cite{huang2018attention}, we divided the 36 and 14 signers into training sets and test sets.

\noindent\textbf{Implementation Details.} 
The encoder and decoder each consist of 3 Sign-GCN layers, while the denoiser has 2 Sign-GCN layers. In the encoder, the input-output channels of Sign-GCNs are 8-16-64-128. In the denoiser, the channels are 32-64-128. The decoder is symmetric to the encoder, with the dimensional transformations in reverse order. In addition, we set the timesteps $t$ of the diffusion model to 1,000 and the number of inferences $i$ to 5. During training, we use the Adam optimizer \cite{Kingma_Ba_2014} and a learning rate of $1 \times 10^{-3}$. Experiments are conducted using PyTorch on NVIDIA GeForce RTX 2080 Ti GPUs.

\noindent\textbf{Evaluation Metrics.}
Due to the current lack of datasets for sign language transition, following prior works~\cite{tang2024discrete}, we simulate realistic discrete sign segments by implementing an interval sampling strategy (remove $Y$ frames every $X$ frames) on large-scale continuous sentence datasets (\emph{i.e.}, PHOENIX14T, USTC-CSL100). We observe that the duration of commenting on sign language is nearly 20 frames. So in our experiments, we mainly adopt the method of removing 20 frames every 30 frames (Remove 20 frames - Every 30 frames) and the method of removing 10 frames every 30 frames (Remove 10 frames - Every 30 frames) to evaluate our performance. Furthermore, we use the isolated sign language dataset (\emph{i.e.}, USTC-SLR500) to assemble discrete vocabulary videos into continuous sequences, demonstrating the model's capability to produce smooth transitions in a real-world setting. We assess the generated transitions from two aspects: \textit{\textbf{semantic accuracy}} and \textit{\textbf{temporal coherence}}. To evaluate the semantic accuracy of generated frames, we use NSLT ~\cite{Camgoz_Hadfield_Koller_Ney_Bowden_2018} to convert generated sign language videos back into text, following prior works~\cite{huang2021towards, saunders2020progressive, saunders2021mixed} in SLP. Then we compare this text with reference data to calculate semantic accuracy metrics including BLEU ~\cite{papineni2002bleu}, ROUGE ~\cite{lin2004rouge}, and WER, while we use DTW to evaluate the temporal coherence of results.


\subsection{Ablation Study}
In this subsection, we explore the effectiveness of Sign-GCN, Multi-scale TCN and related parameters. The experiments are conducted under the condition of removing 20 frames out of every 30 frames (Remove 20 frames - Every 30 frames) to evaluate the semantic accuracy and motion coherence.

\noindent\textbf{Effectiveness of Sign-GCN.}
Table~\ref{tab: ablation-Sign-D2C} shows the ablation results of Sign-GCN module. We use Sign-D2C~\cite{tang2024discrete} as the baseline model, denoted as Base. Base+E-D refers to using Sign-GCN instead of the encoder-decoder architecture in Base, which increases BLEU-1/BLEU-4 by 0.38\%/0.28\% on the TEST set, respectively. When Sign-GCN is used to replace both Encoder-Decoder and Denoiser, the performance is further improved, with BLEU-4 reaching 7.33\% and 6.98\% on the DEV set and TEST set, respectively . This indicates that Sign-GCN preserves richer and more accurate semantics.


\begin{table}[tbp]
\renewcommand\arraystretch{1.0}
\caption{Ablation results of modules on PHOENIX14T.}
\vspace{-3mm}
\label{tab: ablation-Sign-D2C}
\centering
\resizebox{1.0\linewidth}{!}{
\begin{tabular}{cccccccc}
\toprule[1pt]
   \multirow{2}{*}{Methods} & \multicolumn{3}{c}{DEV} & ~ & \multicolumn{3}{c}{TEST} \\
   \cline{2-4}\cline{6-8}
    & B1{$\uparrow$} & B4{$\uparrow$} & ROUGE{$\uparrow$} & ~ & B1{$\uparrow$} & B4{$\uparrow$} & ROUGE{$\uparrow$}\\
\toprule[0.5pt]
   \multicolumn{1}{l}{Base} &19.32 &5.92 &18.33 & ~ &19.59 &6.26 &18.59\\
   \multicolumn{1}{l}{Base+E-D} & 21.12 & 7.16 & 21.71 & ~ &19.97 &6.54 &20.01\\
   \multicolumn{1}{l} {Base+E-D+Denoiser} & \bf{21.55} &\bf{7.33} &\bf{21.87} & ~ &\bf{20.69} &\bf{6.98} &\bf{20.47} \\
\toprule[1pt]
\end{tabular}}
\vspace{-3mm}
\end{table}

\begin{table}[tbp]
\renewcommand\arraystretch{1.0}
\caption{Ablation results of parameters on PHOENIX14T.}
\vspace{-3mm}
\label{tab: ablation-TCN}
\centering
\resizebox{1.0\linewidth}{!}{
\begin{tabular}{ccccc@{\hspace{10pt}}ccc}
\toprule[1pt]
\multirow{2}{*}{Model Config } & \multirow{2}{*}{Module} & \multicolumn{3}{c@{\hspace{10pt}}}{DEV} & \multicolumn{3}{c}{TEST} \\
\cmidrule(lr){3-5} \cmidrule(lr){6-8}
~ & ~ & B1$\uparrow$ & B4$\uparrow$ & DTW$\downarrow$ & B1$\uparrow$ & B4$\uparrow$ & DTW$\downarrow$ \\
\midrule[0.5pt]
7 $\times$ 1 Convolution & TCN & 20.90 & 7.20 & 0.45 & 20.24 & 6.77 & 0.45\\
\textbf{Multi-scale (Ours)} & TCN  & \bf{21.55}&\bf{7.33}&\bf{0.32}&\bf{20.69}&\bf{6.98}&\bf{0.32}\\\hline
w/o residual & Denoiser & 20.73& 6.91 & 0.49& 19.97& 6.54 & 0.49\\
\textbf{w/ residual (Ours)} & Denoiser   &  \bf{21.55}&\bf{7.33}&\bf{0.32}&\bf{20.69}&\bf{6.98}&\bf{0.32}\\
\toprule[1pt]
\end{tabular}}
\vspace{-3mm}
\end{table}

\begin{table}[tbp]
\renewcommand\arraystretch{1.0}
\caption{Ablation results of parameters on PHOENIX14T.}
\vspace{-3mm}
\label{tab: ablation-params}
\centering
\resizebox{1.0\linewidth}{!}{
\begin{tabular}{ccccc@{\hspace{10pt}}ccc}
\toprule[1pt]
\multirow{2}{*}{Param} & \multirow{2}{*}{Value} & \multicolumn{3}{c@{\hspace{10pt}}}{DEV} & \multicolumn{3}{c}{TEST} \\
\cmidrule(lr){3-5} \cmidrule(lr){6-8}
~ & ~ & B1$\uparrow$ & B4$\uparrow$ & ROUGE$\uparrow$ & B1$\uparrow$ & B4$\uparrow$ & ROUGE$\uparrow$ \\
\midrule[0.5pt]
\multirow{4}{*}{k} 
  & 3 & 20.98 & 7.18 & 21.52 & 19.94 & 6.70 & 19.98 \\
  & 5 & 20.93 & 7.12 & 21.71 & 20.52 & 6.82 & \textbf{20.88} \\
  & 7 & \textbf{21.55} & \textbf{7.33} & \textbf{21.87} & \textbf{20.69} & \textbf{6.98} & 20.47 \\
  & 9 & 20.67 & 6.93 & 21.47 & 20.24 & 6.62 & 20.45 \\
\midrule[0.5pt]
\multirow{4}{*}{d} 
  & 1 & 20.75 & 7.05 & 21.51 & 20.12 & 6.73 & 20.18 \\
  & 2 & \textbf{21.55} & \textbf{7.33} & \textbf{21.87} & \textbf{20.69} & \textbf{6.98} & \textbf{20.47} \\
  & 3 & 20.81 & 6.93 & 21.56 & 20.00 & 6.76 & 20.03 \\
  & 4 & 21.16 & 7.26 & 21.76 & 20.03 & 6.64 & 20.02 \\
  \midrule[0.5pt]
\multirow{4}{*}{r} 
  & 0.1 & 21.15 & 7.19 & 21.67 & 19.75 & 6.67 & 19.66 \\
  & 0.4 & 21.19 & 7.14 & 21.66 & 19.77 & 6.66 & 19.68 \\
  & 0.5 & \textbf{21.55} & \textbf{7.33} & \textbf{21.87} & \textbf{20.69} & \textbf{6.98} & \textbf{20.47}\\
  & 0.6 & 21.14 & 7.10 & 21.52 & 20.20 & 6.90 & 20.08 \\
\bottomrule[1pt]
\end{tabular}}
\vspace{-3mm}
\end{table}

\begin{table*}[!htbp]
\renewcommand\arraystretch{1.0}   
\caption{Performance comparison on PHOENIX14T (under setting: Remove 10 frames - Every 30 frames).}
\vspace{-2mm}
\centering
\resizebox{0.95\textwidth}{!}{
\begin{tabular}{lccccccccccccccc}
\toprule[1pt]
\multirow{2}{*}{Methods}&\multicolumn{5}{c}{DEV} & ~ &\multicolumn{5}{c}{TEST} \\
\cline{2-6}
\cline{8-12}
~ &B1$\uparrow$ &B4$\uparrow$ &ROUGE$\uparrow$ &WER$\downarrow$ &DTW$\downarrow$ &  &B1$\uparrow$ &B4$\uparrow$ &ROUGE$\uparrow$ &WER$\downarrow$ &DTW$\downarrow$ \\ 
\midrule[0.5pt]
\multicolumn{1}{l}{Ground Truth}  &29.77 &12.13 &29.60 &74.17&0.00&~ &29.76 &11.93 &28.98 &71.94 &0.00\\ 
\midrule[0.5pt]
\multicolumn{1}{l}{G2P-DDM~\cite{xie2024g2p}} &8.25 &0.99 &7.17 &98.93 &12.71 &~ &9.52 &1.29 &8.12 &98.54 &12.57  \\
\multicolumn{1}{l}{VQ-GCDM~\cite{tang2024GCDM}}  &9.31 &0.87 &7.59 &98.69 &12.11& ~ &9.85 &0.96 &7.95 &98.59 &11.98  \\
\multicolumn{1}{l}{Sign-D2C \cite{tang2024discrete}}
&22.46&8.06&22.07 &80.73&0.62& ~ &22.36&\textbf{8.73}&22.36&79.06 &0.61\\ 
\midrule[0.5pt]
\multicolumn{1}{l}{\textbf{Ours}}
&\bf{23.74}&\bf{8.51}&\bf{24.15} &\bf{77.69}&\bf{0.32} & ~ &\bf{22.64}&8.48&\bf{23.08}&\bf{76.59} &\bf{0.32} \\ 
\bottomrule[1pt]
\end{tabular}}
\label{tab: main1}
\end{table*}

\begin{table*}[!htbp]
\renewcommand\arraystretch{1.0}   
\caption{Performance comparison on PHOENIX14T (under setting: Remove 20 frames - Every 30 frames).}
\vspace{-2mm}
\centering
\resizebox{0.95\textwidth}{!}{
\begin{tabular}{lccccccccccccccc}
\toprule[1pt]
\multirow{2}{*}{Methods}&\multicolumn{5}{c}{DEV} & ~ &\multicolumn{5}{c}{TEST} \\
\cline{2-6}
\cline{8-12}
~ &B1$\uparrow$ &B4$\uparrow$ &ROUGE$\uparrow$ &WER$\downarrow$ &DTW$\downarrow$ &  &B1$\uparrow$ &B4$\uparrow$ &ROUGE$\uparrow$ &WER$\downarrow$ &DTW$\downarrow$ \\ 
\midrule[0.5pt]
\multicolumn{1}{l}{Ground Truth}  &29.77 &12.13 &29.60 &74.17&0.00&~ &29.76 &11.93 &28.98 &71.94 &0.00\\ 
\midrule[0.5pt]
\multicolumn{1}{l}{G2P-DDM~\cite{xie2024g2p}} &8.98 &0.63 &7.44 &98.61 &15.95  &~ &9.13 &0.74 &7.75 &98.64 &15.58  \\
\multicolumn{1}{l}{VQ-GCDM~\cite{tang2024GCDM}} &8.64 &0.89 &7.08 &98.48 &12.40 & ~ &9.02 &1.36 &7.22 &98.61 &12.22 \\
\multicolumn{1}{l}{Sign-D2C \cite{tang2024discrete}}
&19.32&5.92&18.33 &87.30&0.70 & ~ &19.59&6.26&18.59&86.80 &0.68 \\ 
\midrule[0.5pt]
\multicolumn{1}{l}{\textbf{Ours}}
&\bf{21.55}&\bf{7.33}&\bf{21.87} &\bf{85.21}&\bf{0.32} & ~ &\bf{20.69}&\bf{6.98}&\bf{20.47}&\bf{84.50} &\bf{0.32} \\ 
\bottomrule[1pt]
\end{tabular}}
\label{tab: main2}
\end{table*}

\noindent\textbf{Effectiveness of Multi-scale Temporal Modeling.}
To evaluate the impact of multi-scale temporal feature extraction, we compare our proposed Multi-scale TCN with a standard $7 \times 1$ convolution-based TCN under a controlled setting. As shown in Table~\ref{tab: ablation-TCN}, our Multi-scale TCN yields significant improvements across all metrics on both the DEV and TEST sets. Specifically, on the TEST set, Multi-scale TCN surpasses the baseline by +0.45\% BLEU-1, +0.21\% BLEU-4 and -0.13\% DTW. This highlights Multi-scale TCN's ability to pick up complex temporal correlations that promotes smooth and accurate flow in generated sign videos.

\noindent\textbf{Residual Connections in Denoiser.}
We further investigate the role of residual connections within the denoising module as shown in Table~\ref{tab: ablation-TCN} . Removing residual connections (w/o residual) results in a noticeable performance drop, with the TEST DTW score increasing from 0.32\% to 0.49\%. Similarly, both BLEU-1 and BLEU-4 metrics decline. The residual-enhanced variant (w/ residual) achieves the  improvement of overall performance. These results strongly demonstrate that residual connections of condition information can preserve more semantic relevance for our model.

\begin{table}
\renewcommand\arraystretch{1.0}   
\caption{Performance comparison on USTC-CSL100 (under setting: Remove 20 frames - Every 30 frames).}
\vspace{-1mm}
\centering
\resizebox{1.0\columnwidth}{!}{
\begin{tabular}{lccccccc}
\toprule[1pt]
Methods & B1$\uparrow$ & B4$\uparrow$ & ROUGE$\uparrow$ & WER$\downarrow$ & DTW$\downarrow$  \\
\midrule[0.5pt]
Ground Truth & 69.10 & 59.12 & 68.53 & 47.38 & 0.00  \\
\midrule[0.5pt]
G2P-DDM~\cite{xie2024g2p} & 13.43 & 1.74 & 13.62 & 92.15 & 7.47  \\
VQ-GCDM~\cite{tang2024GCDM} & 18.77 & 2.08 & 16.64 & 92.67 & 10.89\\
Sign-G2C \cite{tang2024discrete}& 67.95 & 56.61 & \textbf{67.38} & \textbf{49.24}& 1.79  \\
\midrule[0.5pt]
\textbf{Ours}   & \textbf{68.07} & \textbf{57.11} & 67.15 & 49.38 & \textbf{1.40} \\
\bottomrule[1pt]
\end{tabular}}
\label{tab:csl100_half}
\end{table}

\noindent\textbf{Exploration of parameters.}
Table ~\ref{tab: ablation-params} compares various Sign-GCN settings, including different values of  $k$, $d$, where $k$ represents the temporal kernel size, and $d$ represents the dilation rate for the multi-scale TCN operations. We first observe that all configurations consistently outperform the baseline, confirming the stability of Sign-GCN as a spatial-temporal feature extractor. Specifically, increasing the kernel size $k$ from 3 to 7 leads to a consistent improvement across all metrics, with the best performance observed at $k$=7, achieving BLEU-1 scores of 21.55\% on DEV and 20.69\% on TEST. However, further increasing $k$ to 9 results in a slight drop in performance. As for the dilation rate $d$, we find that the performance peaks at $d$=2, with ROUGE scores of 21.87\% on DEV and 20.47\% on TEST.  Both lower dilation and higher dilation lead to marginal performance drops. This suggests that moderate dilation enables a better balance between capturing local dependencies and maintaining global contextual awareness. These results confirm that $k$=7 and $d$=2 is an effective configuration for Sign-GCN. Meanwhile, table \ref{tab: ablation-params} also shows masking ratio r = 0.5 is a good balance between semantic retention and transition smoothness.

 \begin{figure*}[tbh]
    \centering
    \includegraphics[width=1.0\textwidth]{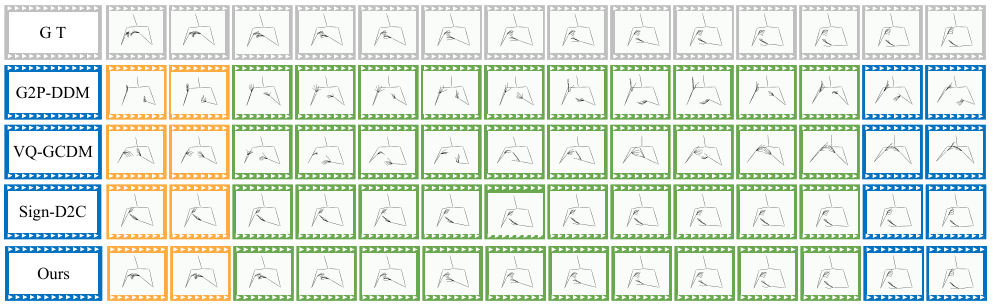}
    \caption{Visualization examples of generating transition pose under setting "Remove 10 frames - Every 30 frames" on PHOENIX14T. We compare our method with three SOTA methods, attached the Ground Truth. \textcolor{green}{Green}: Transitions. \textcolor{orange}{Orange} and \textcolor{blue}{Blue}: Observations.}
    \label{ablation}
\end{figure*}

\begin{figure*}[tbh]
    \centering
    \includegraphics[width=1.0\textwidth]{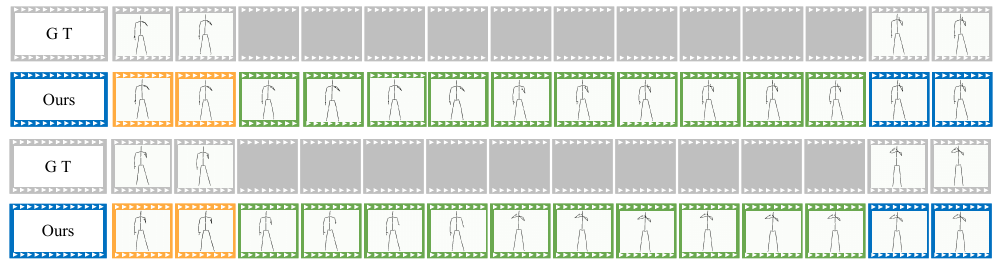}
    \caption{Visualization examples on the USTC-SLR500, demonstrate the generation of transition poses in a realistic word-level scenario. \textcolor{green}{Green}: Transitions.\textcolor{orange}{Orange} and \textcolor{blue}{Blue}: Observations.}
    \label{slr 500}
\end{figure*}

\subsection{Comparison with State-of-the-Arts}
Due to the current lack of benchmarks for sign language transition generation, we adapted the SLP models (\emph{i.e.} G2P-DDM~\cite{xie2024g2p} and VQ-GCDM~\cite{tang2024GCDM}) as benchmarks. Additionally, we compare with the sign language transition method (\emph{i.e.} Sign-D2C~\cite{tang2024discrete}). 

\noindent\textbf{PHOENIX14T} Table \ref{tab: main1} shows the results under the "Remove 10 frames - Every 30 frames" setting, StgcDiff significantly outperforms two compared models G2P-DDM and VQ-GCDM, achieving 23.74\% and 22.64\% BLEU-1 on the DEV and TEST sets, respectively. Even compared with the best-performing method Sign-D2C, our method achieves better performance on most metrics. Furthermore, our method shows significant gains in temporal coherence, \emph{e.g.}, StgcDiff is 0.30\% and 0.29\% lower DTW than Sign-D2C \cite{tang2024discrete} on DEV and TEST. These results confirm that our model generates smoother and more realistic transitions than the baseline methods.

To further prove the performance of our method, we compare with other methods under the setting "Remove 20 frames - Every 30 frames" as shown in Table \ref{tab: main2}. In this challenging setting, our method achieves superior performance on the all metrics. We achieve BLEU-1/BLEU-4 scores of 21.55\%/7.33\% on DEV and 20.69\%/6.98\% on TEST. Meanwhile, our method obtains higher ROUGE and lower WER scores, demonstrating the better performance in semantic relevance. Moreover, our method is especially significantly better than Sign-D2C by DTW score of 0.32 on the DEV and TEST sets, further highlighting the advantages of our approach in accurately generating natural transition poses. 

\noindent\textbf{USTC-CSL100}
Table \ref{tab:csl100_half} shows comparison results on a challenging Chinese sign language benchmark USTC-CSL under the more challenging setting "Remove 20 frames - Every 30 frames". Our proposed method achieves competitive performance, with a BLEU-1 score of 68.07\% and a BLEU-4 score of 57.11\%, which are higher than other baseline methods. In terms of ROUGE, our method performs similarly to the best methods, reaching 67.15\%. For WER, our model also shows a notable result with a score of 49.38\%, which is close to the performance of  the best-performing method Sign-D2C. Regarding DTW, our method shows the best results with values of 1.40, indicating better alignment and  more coherent transitions compared to other methods. 

\subsection{Qualitative Results}
\noindent\textbf{Visualization Results.}
In Figure \ref{ablation}, we visualize transition sign poses generated by the proposed StgcDiff and other methods, $i.e.$, G2P-DDM \cite{xie2024g2p}, VQ-GCDM \cite{tang2024GCDM} and Sign-D2C \cite{tang2024discrete} . This example clearly shows the transition frames generated by StgcDiff are noticeably superior to another three methods. Specifically, our method significantly outperforms G2P-DDM and VQ-GCDM in terms of temporal coherence, as the latter two exhibit unnatural abruption. In comparison of Sign-D2C, our method has an advantage in spatial modeling of skeleton joints, further highlighting the good spatial representation learning ability of our proposed model.

\noindent\textbf{Verification in Realistic Scenarios.}
Figure \ref{slr 500} presents two visual examples illustrating the synthesis of continuous sign language videos from discrete segments in a realistic word-level context, based on the USTC-SLR500 dataset. In each example, the observed frames from the two separate sign segments are highlighted in blue and orange, while the generated transition frames are depicted in green. The visualizations indicate that our approach produces transitions that are not only temporally smooth but also spatially consistent, effectively bridging the gap between isolated segments and yielding a seamless, coherent sign language video.

\section{Conclusions}
In this work, we propose a novel graph-based diffusion framework for sign transition generation, effectively capturing spatial-temporal dependencies to produce smooth and contextually accurate transitions. The designed Sign-GCN module models both the spatial relationships between joints and the multi-scale temporal dependencies in sign language sequences. Extensive experiments on three benchmarks validate the superior performance of the proposed method. 


\bibliographystyle{ACM-Reference-Format}
\bibliography{sample-base}

\appendix

\end{document}